\documentclass[10pt,conference]{IEEEtran}
\IEEEoverridecommandlockouts
\usepackage{cite}
\usepackage{booktabs} % 用于三线表 \toprule, \midrule, \bottomrule
\usepackage{amsmath,amssymb,amsfonts}
\usepackage{algorithmic}
\usepackage{graphicx}
\usepackage{textcomp}
\usepackage{xcolor}
\usepackage[hidelinks]{hyperref}
\usepackage{multirow}
\usepackage{arydshln} % 用于虚线 \hdashline (如果报错请删去此包并将 \hdashline 改为 \midrule)
\usepackage{bm}
\IEEEoverridecommandlockouts

\def\BibTeX{{\rm B\kern-.05em{\sc i\kern-.025em b}\kern-.08em
    T\kern-.1667em\lower.7ex\hbox{E}\kern-.125emX}}
\begin{document}

\title{Graph-RHO: Critical-path-aware Heterogeneous Graph Network for Long-Horizon Flexible Job-Shop Scheduling
}

\author{\IEEEauthorblockN{Yujie Li}
\IEEEauthorblockA{\textit{Beijing University of} \\\textit{Posts and Telecommunications} \\
Beijing, China \\
liyujie2003@bupt.edu.cn}
\and
\IEEEauthorblockN{Jiuniu Wang}
\IEEEauthorblockA{\textit{City University of Hong Kong} \\
Hong Kong, China \\
wangjiuniu@gmail.com}
\and
\IEEEauthorblockN{Mugen Peng}
\IEEEauthorblockA{\textit{Beijing University of} \\\textit{Posts and Telecommunications} \\
Beijing, China \\
pmg@bupt.edu.cn}
\and
\IEEEauthorblockN{Guangzuo Li}
\IEEEauthorblockA{\textit{Chinese Academy of Sciences} \\
Beijing, China \\
ligz@aircas.ac.cn}
\and
\IEEEauthorblockN{Wenjia Xu}
\IEEEauthorblockA{\textit{Beijing University of} \\\textit{Posts and Telecommunications} \\
Beijing, China \\
xuwenjia@bupt.edu.cn}
\thanks{This work has been funded by the National Natural Science Foundation of China under Grant 62301063.}
}

\maketitle

\begin{abstract}
Long-horizon Flexible Job-Shop Scheduling~(FJSP) presents a formidable combinatorial challenge due to complex, interdependent decisions spanning extended time horizons. While learning-based Rolling Horizon Optimization~(RHO) has emerged as a promising paradigm to accelerate solving by identifying and fixing invariant operations, its effectiveness is hindered by the structural complexity of FJSP. Existing methods often fail to capture intricate graph-structured dependencies and ignore the asymmetric costs of prediction errors, in which misclassifying critical-path operations is significantly more detrimental than misclassifying non-critical ones. Furthermore, dynamic shifts in predictive confidence during the rolling process make static pruning thresholds inadequate.
To address these limitations, we propose Graph-RHO, a novel critical-path-aware graph-based RHO framework. First, we introduce a topology-aware heterogeneous graph network that encodes subproblems as operation-machine graphs with multi-relational edges, leveraging edge-feature-aware message passing to predict operation stability. Second, we incorporate a critical-path-aware mechanism that injects inductive biases during training to distinguish highly sensitive bottleneck operations from robust ones. Third, we devise an adaptive thresholding strategy that dynamically calibrates decision boundaries based on online uncertainty estimation to align model predictions with the solver's search space. Extensive experiments on standard benchmarks demonstrate that \mbox{Graph-RHO} establishes a new state of the art in solution quality and computational efficiency. Remarkably, it exhibits exceptional zero-shot generalization, reducing solve time by over 30\% on large-scale instances (2000 operations) while achieving superior solution quality. Our code is available \href{https://github.com/IntelliSensing/Graph-RHO}{here}.

\end{abstract}

\begin{IEEEkeywords}
Graph neural network, flexible job-shop scheduling, multi-task learning.
\end{IEEEkeywords}

\begin{figure}[tb]
    \centering
    \includegraphics[width=0.95\linewidth]{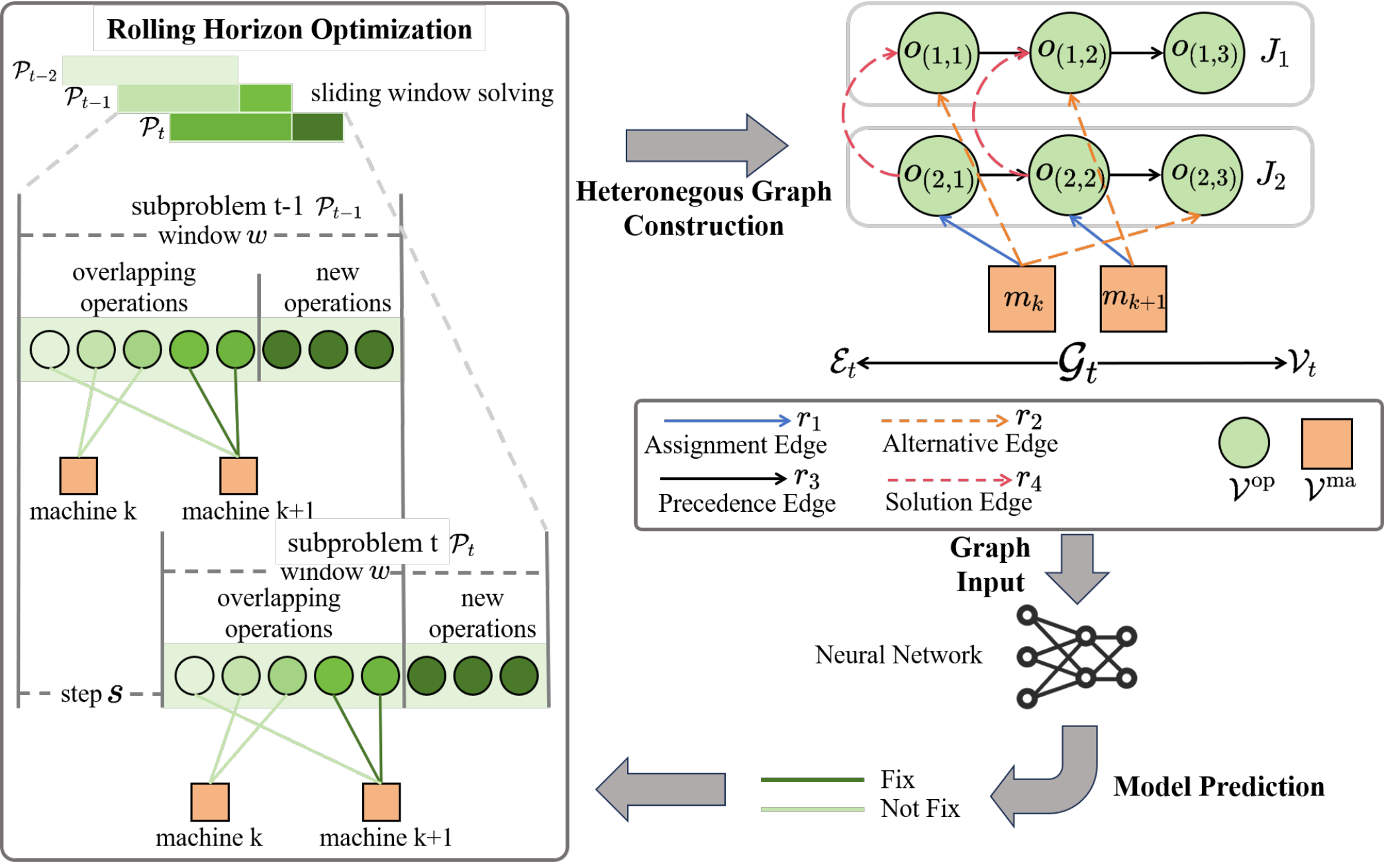}
    \caption{\textbf{Illustration of Graph-RHO.} Within the RHO loop, each subproblem is mapped to a heterogeneous graph that explicitly encodes topological constraints via multi-relational edges. Leveraging this structural representation, the neural network identifies stable operations~(solid dark green lines) and fixes their machine assignments.}
    \label{fig:teaser}
\end{figure}

\section{Introduction}
The Flexible Job-Shop Scheduling Problem~(FJSP) is a cornerstone of modern smart manufacturing and logistics, governing the efficiency of resource allocation in complex production systems~\cite{thames2016software}. In FJSP, jobs composed of sequentially constrained operations must be scheduled onto a candidate machine pool, requiring simultaneous decisions on machine assignment and execution order for every operation. In industrial practice, these problems manifest as long-horizon tasks, necessitating the optimization of thousands of operations with intricate precedence constraints over extended time horizons. While exact solvers and meta-heuristics~\cite{google_cpsat, li2016effective} can handle small-scale instances, they suffer from the ``curse of dimensionality'' in long-horizon settings, where the combinatorial search space expands exponentially, rendering global optimization computationally intractable.

To tackle this scalability challenge, Rolling Horizon Optimization~(RHO)~\cite{glomb2022rolling, mattingley2011receding, sethi1991theory} has emerged as a standard paradigm. As shown in Fig.~\ref{fig:teaser}, to reduce global complexity, RHO decomposes the problem into manageable subproblems and solves them iteratively using a sliding window that advances in fixed steps, to cover both overlapping and new operations. 
% However, a critical bottleneck arises because traditional RHO redundantly re-optimizes all overlapping operations, disregarding the fact that part of the overlapping operations retain their prior assignments. This significant computational redundancy drastically reduces solving efficiency, rendering standard RHO schemes impractical for real-world long-horizon scenarios where rapid responsiveness is important. 
However, a critical bottleneck arises because conventional RHO repeatedly re-optimizes all overlapping operations, even though a substantial portion of them preserve stable prior operation-machine assignments, leading to significant computational redundancy.
To mitigate this, the recent Learning-based RHO framework L-RHO~\cite{li2025learning} pioneered a data-driven approach that uses an MLP network to identify and fix stable operation assignments, thereby successfully pruning the search space. However, limited by its vector-based representation, the MLP network abstracts interactions among operations and machines into simplified linear dependencies, which inadequately capture the underlying physical constraints and structural properties of FJSP.

Building upon this paradigm, we draw inspiration from the intrinsic graph-structured nature of FJSP, where operation sequences and machine constraints are deeply intertwined, as shown in Fig.~\ref{fig:teaser}. We observe that the stability of overlapping operations is fundamentally governed by these complex topological dependencies rather than simple statistical correlations. This insight motivates the construction of Graph Neural Networks to explicitly model the job shop floor as a relational graph. By interpreting the structural relationships between precedence and resource contention, we can achieve significantly more precise identification and fixation of invariant operations within the rolling windows.
% While this paradigm marks a significant leap in enabling scalable scheduling, the current implementation relies on generic MLPs and static inference strategies, which leaves room for further improvement in complex environments. Specifically, the MLP-based encoder relies on statistical aggregations that are fundamentally misaligned with the intrinsic graph-structured nature of FJSP. Failing to explicitly model the complex dependencies arising from operation precedence and machine contention, the method loses critical topological context, severely limiting its generalization across varying problem scales. Furthermore, the standard training objective treats all operations indiscriminately, potentially overlooking the asymmetric cost of erroneously fixing critical-path operations, where a mistake is far more damaging than on non-critical nodes. Finally, the reliance on static thresholds ignores the dynamic uncertainty of the rolling process, which can lead to either rigid over-pruning or inefficient under-pruning.

Motivated by these, we propose Graph-RHO, a novel, critical-path-aware \textbf{Graph}-based \textbf{RHO} framework that advances the learning-based decomposition paradigm. Specifically, we introduce a topology-aware heterogeneous graph encoder to align the representation mechanism with the intrinsic graph-structured nature of FJSP. By explicitly modeling the shop floor via edge-feature-aware message passing, this module captures the constraint patterns embedded in the topology, enabling the model to reason about complex dependencies rather than merely fitting statistical correlations. Moreover, we introduce a critical-path-aware mechanism that incorporates a critical path identification objective during training. This auxiliary task injects an inductive bias, forcing the model to distinguish highly sensitive bottleneck operations from robust ones, thereby ensuring robust search-space pruning. Furthermore, we devise an adaptive thresholding inference strategy that dynamically aligns the pruning threshold with the model's online uncertainty distribution, creating a deep synergy between the neural predictor and the combinatorial solver. 

% To validate our framework, we conduct rigorous experiments across diverse long-horizon FJSP settings, benchmarking against a comprehensive suite of exact, meta-heuristic, and learning-based baselines. Empirical results confirm that Graph-RHO achieves state-of-the-art performance, striking a superior balance between efficiency and solution quality. Furthermore, our model exhibits exceptional zero-shot generalization on unseen large-scale instances, and extensive ablation studies verify the essential contributions of the proposed topological encoder, the critical-path-aware mechanism, and the adaptive thresholding strategy. \wenjia{This paragraph is redundancy. Remove it and merge it with the contribution 4.}

Our main contributions are summarized as follows:
\begin{itemize}
    \item We propose Graph-RHO, a learning-based RHO framework that leverages a heterogeneous graph encoder to explicitly model the FJSP shop floor. Through edge-feature-aware message passing, this module captures multi-relational topological constraints, significantly enhancing the representation capability for complex scheduling dynamics.
    \item We introduce a critical-path-aware mechanism. By incorporating an auxiliary critical path identification training objective, we inject an inductive bias that promotes the model to distinguish high-sensitivity critical operations from robust ones, thereby safeguarding solution quality against erroneous pruning.
    \item We devise an adaptive thresholding inference strategy. By dynamically calibrating decision boundaries based on online uncertainty estimation, this mechanism establishes a self-adjusting synergy between the neural predictor and the combinatorial solver, effectively balancing pruning efficiency with feasibility.
    \item Extensive benchmarking against a comprehensive suite of exact, meta-heuristic, and learning-based baselines confirms that Graph-RHO establishes a new state-of-the-art across various long-horizon FJSP settings. Crucially, the model demonstrates exceptional zero-shot generalization in both scale expansion and load intensification scenarios. Our model maintains superior solving efficiency and solution quality, effectively overcoming distribution shifts.
\end{itemize}

\begin{figure*}[tb]
    \centering
    \includegraphics[width=1\linewidth]{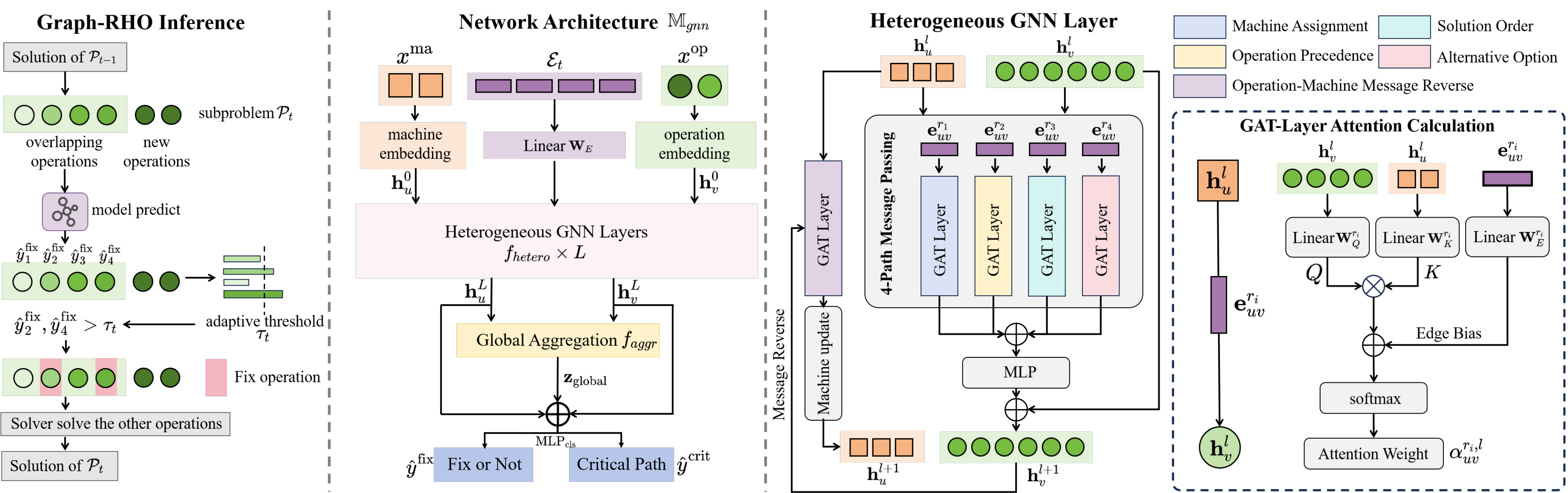}
    \caption{\textbf{Overview of Graph-RHO:} \textbf{i)} \textbf{Graph-RHO Inference}: The model predicts stability probabilities for overlapping operations, employing an adaptive threshold $\tau_t$ to fix high-confidence variables and prune the search space for subproblem $\mathcal{P}_t$. \textbf{ii)} \textbf{Network Architecture}: The $\mathbb{M}_{gnn}$ processes heterogeneous graph states via stacked GNN layers and global aggregation~($f_{aggr}$) to simultaneously optimize operation stability~($\hat{y}^{\text{fix}}$) and path criticality~($\hat{y}^{\text{crit}}$) predictions. \textbf{iii)} \textbf{Heterogeneous GNN Layer}: The layer executes 4-path message passing to capture multi-relational dependencies, coupled with a reverse flow for machine updates. The \textbf{GAT Inset} illustrates the edge-feature-aware attention mechanism, where edge constraints $\mathbf{e}_{uv}^{r_i}$ are injected as structural bias.}
    \label{fig:framework}
\end{figure*}

\section{Related Works}
\subsection{Long Horizon FJSP}
Traditional FJSP solvers, including exact methods~\cite{google_cpsat} and meta-heuristics~\cite{li2016effective}, struggle with the exponential time complexity of long-horizon scenarios. While Deep Reinforcement Learning~(DRL) offers promise, constructive agents~\cite{song2022flexible, zhang2023dynamic, huang2025deep} often fail to scale effectively. To address this, Rolling Horizon Optimization~(RHO)~\cite{garcia1989model, lu2022train} decomposes the problem into iterative overlapping subproblems. However, standard RHO suffers from computational redundancy by repeatedly re-optimizing invariant variables. The recent L-RHO framework~\cite{li2025learning} mitigates this by learning to fix stable operations within overlapping subproblem windows. Yet, L-RHO relies on topology-agnostic MLPs, failing to capture intrinsic graph constraints, which highlights the need for structurally aligned representations.

\subsection{GNNs in Combinatorial Optimization}
Graph Neural Networks (GNNs) have proven effective at capturing topological structures for Combinatorial Optimization. In scheduling, GNNs are widely used to encode disjunctive graphs for DRL-based dispatching~\cite{song2022flexible, teichteil2023fast} or to enhance exact solvers via Learning-to-Branch~\cite{wang2024learning, labassi2022learning}. Despite these advances, the integration of GNNs into RHO remains unexplored. Existing learning-based RHO methods rely on MLPs, neglecting critical edge features like precedence and resource contention. Our work bridges this gap by introducing a heterogeneous graph network tailored for FJSP RHO, shifting the paradigm from statistical fitting to topological reasoning.

\section{Methodology}
In this section, we present the design of our Graph-RHO, a learning-based RHO framework that accelerates long-horizon FJSP. We first formalize the Flexible Job-Shop Scheduling Problem and the Rolling Horizon Optimization paradigm in section~\ref{Problem Formulation}. Then detail the three pillars of our proposed method: the heterogeneous graph network $\mathbb{M}_{gnn}$ in Sec.~\ref{GNN}, the critical-path-aware mechanism $\mathbb{M}_{cpa}$ in Sec.~\ref{MTL}, and finally, the adaptive thresholding strategy $\mathbb{M}_{thr}$ for inference in Sec.~\ref{Adaptive}.

\subsection{Problem Formulation}\label{Problem Formulation}
We model the Flexible Job-Shop Scheduling Problem~(FJSP) as a discrete optimization problem defined by a tuple $\langle$machines, jobs,  operations$\rangle$ as $\langle \mathcal{M}, \mathcal{J}, \mathcal{O} \rangle$. Here $\mathcal{M} = \{m_1, \dots, m_{N_m}\}$ is a set of $N_m$ machines and $\mathcal{J} = \{J_1, \dots, J_{N_j}\}$ is a set of $N_j$ jobs. Each job $J_i$ is defined by an ordered sequence of operations $(o_{(i,1)}, \dots, o_{(i, n_i)}) \in \mathcal{O}$, governed by the linear precedence constraints $o_{(i,k)} \rightarrow o_{(i,k+1)}$ for $k = 1, \dots, n_i-1$. For each operation $o_{(i,k)}$, the solver must make two interdependent decisions: \textbf{i)} Select a machine $m_{(i,k)} \in M_{(i,k)} \subseteq \mathcal{M}$ with deterministic processing time $p_{i,k}$; \textbf{ii)} Determine the start time $s_{i,k} \ge 0$ of each operation. The optimization goal is to find a valid schedule $\Pi = \{(m_{(i,k)}, s_{i,k}) \mid \forall o_{i,k} \in \mathcal{O}\}$ that minimizes the makespan $C_{\max} = \max\limits_{i,k} (s_{i,k} + p_{i,k})$, which corresponds to the maximum completion time over all operations, subject to precedence constraints and machine capacity constraints.

To overcome the computational intractability of global optimization, Rolling Horizon Optimization (RHO) employs an iterative sliding-window mechanism. At each iteration $t$, RHO constructs a subproblem $\mathcal{P}_t$ containing $w$ operations. Solving $\mathcal{P}_t$ yields a local schedule $\Pi_t$, from which only the immediate $s$ operations are committed. The remaining $w-s$ operations form the overlap region, denoted as $\mathcal{O}_{t}^{overlap}$, which is traditionally deferred for full re-optimization. To mitigate the redundancy of repeatedly optimizing this region, the learning-based paradigm exploits the stability hypothesis~\cite{li2025learning}, which posits that a subset of operations $\mathcal{O}_{t}^{fix} \subseteq \mathcal{O}_{t}^{overlap}$ retains invariant machine assignments across consecutive windows. By utilizing a model to predict and fix the variables in $\mathcal{O}_{t}^{fix}$ to their values from $\Pi_{t-1}$, the solver's search space for the next iteration is effectively pruned. To ensure precise identification of $\mathcal{O}_{t}^{fix}$, Graph-RHO employs a heterogeneous graph network to extract topological dependencies. Specifically, during inference~(see in Fig.~\ref{fig:framework}), the model predicts fixation scores for overlapping operations and applies an adaptive threshold $\tau_t$ to dynamically distinguish the most reliable subset, which are fixed to their assignments from $\Pi_{t-1}$, thereby pruning the search space for the solver.

\subsection{Graph-GHO Framework}
We propose the critical-path-aware heterogeneous \textbf{Graph}-based \textbf{R}olling \textbf{H}orizon \textbf{O}ptimization~(Graph-RHO) framework. Graph-RHO synergizes three core advancements to identify stable operations: \textbf{i)} a topology-aware heterogeneous graph network $\mathbb{M}_{gnn}$ that explicitly encodes the intrinsic disjunctive graph structure via edge-feature-aware message passing; \textbf{ii)} a critical-path-aware mechanism $\mathbb{M}_{cpa}$ that incorporates critical path identification as an auxiliary task during training to rectify the sensitivity-agnostic limitation where standard classification treats all operations equally; and \textbf{iii)} a neural-symbolic synergy adaptive thresholding inference strategy $\mathbb{M}_{thr}$ that calibrates decision boundaries against predictive uncertainty shifts.

\subsection{Heterogeneous Graph Network}\label{GNN}

Effective resolution of FJSP demands a precise understanding of the intricate topological associations between operation and machine nodes. To this end, we construct a heterogeneous graph network, denoted as $\mathbb{M}_{gnn}$~(Fig.~\ref{fig:framework}). Structurally, this model comprises stacked heterogeneous graph neural network layers $f_{hetero}$ for constraint propagation, a global aggregation module $f_{aggr}$ for system-level context integration, and a dedicated task head $\text{MLP}_{\text{cls}}$. In the following, we first detail the construction of the heterogeneous graph that encodes the subproblem state. Next, we introduce the stacked heterogeneous graph neural network layers $f_{hetero}$. Finally, we describe the global aggregation module $f_{aggr}$ and task head for the ultimate prediction.

\textbf{Heterogeneous Graph.} At each iteration $t$, we map the state of the current subproblem $\mathcal{P}_t$ to a heterogeneous graph $\mathcal{G}_t = (\mathcal{V}_t, \mathcal{E}_t)$ indicating $($nodes, edges$)$, as illustrated in Fig.~\ref{fig:teaser}. In the following, we detail the construction of node features and heterogeneous edges to capture the system state.

The node set $\mathcal{V}_t = \mathcal{V}^{\text{op}} \cup \mathcal{V}^{\text{ma}}$ comprises two distinct types of entities, defined as follows: \textbf{i)} operation nodes $\mathcal{V}^{\text{op}}$: For each planed operation $o_{(i,k)}$, we construct a feature vector $x^{\text{op}} \in \mathbb{R}^{15}$ containing static attributes, such as processing times and job IDs, alongside dynamic states like the current start time $s_{i,k}$ and overlap status. \textbf{ii)} machine nodes $\mathcal{V}^{\text{ma}}$: For each machine $m_j \in \mathcal{M}$, we construct a feature vector $x^{\text{ma}} \in \mathbb{R}^{11}$ summarizing its workload statistics, including the average completion time and the number of assigned overlapping operations.

To decouple the complex constraints in FJSP, the edge set $\mathcal{E}_t$ consists of a set of relations $\mathcal{R}$ containing four distinct semantic edge types $\{r_1,r_2,r_3,r_4\}$. For an operation node $v$ and its neighbor $u$, an edge $e_{uv}^{r_i} \in \mathcal{E}_t$ of relation type $r_i \in \mathcal{R}$ carries an edge feature vector $\mathbf{e}_{uv}^{r_i}$. These relations are defined as: \textbf{i)} machine assignment $r_1$ ($\mathcal{O} \to \mathcal{M}$): This represents the current tentative assignment of operation $v$ to machine $u$. The edge feature includes the processing duration $p_{i,k}$ and an assignment indicator to encode direct resource occupation. \textbf{ii)} alternative option $r_2$ ($\mathcal{O} \to \mathcal{M}$): This links operation $v$ to other compatible machines $u' \in \{M_{(i,k)} - \{u\} \}$, enabling the model to perceive the opportunity cost of the current assignment. \textbf{iii)} precedence constraint $r_3$ ($\mathcal{O} \to \mathcal{O}$): These are directed edges from $o_{(i,k-1)}$ to $o_{(i,k)}$ representing job-sequence constraints defined in $\mathcal{J}$. \textbf{iv)} solution order $r_4$ ($\mathcal{O} \to \mathcal{O}$): These directed edges represent the execution sequence on the same machine derived from the previous local schedule $\Pi_{t-1}$.

By unifying these node entities and semantic edges, $\mathbb{M}_{gnn}$ establishes a topology-complete representation that inherently possesses permutation invariance. This design empowers the model to effectively capture the intrinsic topological structure of FJSP. As corroborated by empirical results, this structural alignment endows the model with powerful topological perception capabilities and exceptional zero-shot generalization, enabling the learned policy to seamlessly transfer across varying scales without retraining.

\textbf{Heterogeneous Graph Neural Network Layers.} 
To effectively extract topological relation and simulate constraint propagation, $\mathbb{M}_{gnn}$ stacks $L$ heterogeneous graph neural network layers $f_{hetero}$. Within each layer, we adopt a relation-specific edge-feature-aware message-passing mechanism in which operation nodes aggregate information from neighbors based on their distinct relational types $r_i \in \mathcal{R}$.
%$\mathbb{M}_{gnn}$ employs an iterative message passing mechanism to simulate constraint propagation. By stacking $L$ heterogeneous graph neural network layers $f_{hetero}$, the network captures cascading dependencies bi-directionally, where operations aggregate resource availability from machines while machines aggregate workload demands from operations.

For a target operation, its scheduling status is intrinsically determined by the states of its topological neighbors, such as the current workload of compatible machines or the completion progress of preceding operations. To capture these interactions, we calculate attention coefficients to weigh the importance of each neighbor with graph attention networks~(GATs)~\cite{velivckovic2017graph}. However, standard graph attention is insufficient for scheduling as it neglects the quantitative attributes of the connections. To accurately reflect physical constraints, we explicitly inject edge features $\mathbf{e}_{uv}^{r_i}$ into the attention computation~\cite{vaswani2017attention}. This design acts as a structural bias, ensuring that neighbors with significant attributes, such as longer processing times, exert a stronger influence on the target node.

Formally, let $\mathbf{h}_v^{l}$ and $\mathbf{h}_u^{l}$ denote the embeddings of node $v$ and neighbor $u$ at layer $l$. For a target operation $v$, we compute the constraint-aware attention coefficient $\alpha_{uv}^{r_i, l}$ under relation $r_i$ as follows~(visualized in the \textbf{GAT-Layer Attention Calculation} inset of Fig.~\ref{fig:framework}):
\begin{align}
\text{score}_{uv}^{{r_i}, l} &= \frac{(\mathbf{W}_Q^{r_i} \mathbf{h}_v^{l})^\top (\mathbf{W}_K^{r_i} \mathbf{h}_u^{l})}{\sqrt{d_k}} + \mathbf{W}_E^{r_i} \cdot \mathbf{e}_{uv}^{r_i} \,, \label{eq:score} \\
\alpha_{uv}^{{r_i}, l} &= \text{softmax}_{u \in \mathcal{N}_{r_i}(v)} \left( \text{score}_{uv}^{{r_i}, l} \right) \,, \label{eq:alpha}
\end{align}
where $\mathbf{W}_Q^{r_i}$, $\mathbf{W}_K^{r_i}$ and $\mathbf{W}_E^{r_i}$ are learnable weight matrices, $\mathcal{N}_{r_i}(v)$ is the neighbors of the node $v$, and $\mathbf{W}_E^{r_i}$ projects the edge features into the attention space to modulate the connection weight dynamically.

With the attention coefficients established, operation nodes aggregate messages from all four relation types. These heterogeneous messages are concatenated and fused via a residual MLP to update the operation state $\mathbf{h}_v^{l+1}$:
\begin{align}
\mathbf{m}_v^{{r_i}, l} &= \sum_{u \in \mathcal{N}_{r_i}(v)} \alpha_{uv}^{{r_i}, l} \left( \mathbf{W}_V^{r_i} \mathbf{h}_u^{l} \right) \,, \label{eq: m} \\
\mathbf{h}_v^{l+1} &= \text{Norm} \left( \mathbf{h}_v^{l} + \text{MLP} \left( \mathop{\big\|_{r_i \in \mathcal{R}}} \mathbf{m}_v^{{r}_i, l} \right) \right) \,, \label{eq: h}
\end{align}
where $\big\|$ denotes concatenation.

Crucially, we introduce a reverse message passing step to complete the constraint loop. Machine nodes aggregate messages from their assigned operations via reversed assignment edges to update their own embeddings $\mathbf{h}_u^{l+1}$. This mechanism allows machines to dynamically reflect current congestion levels and broadcast this resource availability back to operations in the subsequent layer. By grounding predictions in this closed-loop structural causality, the model achieves robust reasoning about resource contention and precedence states within the local receptive field.

\textbf{Prediction.} 
After $L$ message passing $f_{hetero}$ layers, we obtain final embeddings for all nodes. Before the final prediction, we perform a global aggregation step in $f_{aggr}$. We apply average pooling across all operations and machine nodes to obtain a global context vector $\mathbf{z}_{\text{global}}$, which is then concatenated with the node-wise embeddings, ensuring that local decisions are conditioned on the global system load. For each candidate operation $o_{(i,k)} \in \mathcal{O}_{t}^{overlap}$, we construct a final comprehensive representation $\mathbf{h}_{o_{(i,k)}}^{\text{final}}$ by concatenating its local operation embedding, the global context, and the embedding of its currently assigned machine $\mathbf{h}_{u}^{L}$:
\begin{align}
\mathbf{h}_{o_{(i,k)}}^{\text{final}} = \left [ \mathbf{h}_{v}^{L} ; \mathbf{z}_{\text{global}}   ; \mathbf{h}_{u}^{L}\right ] \,,
\end{align} where $\left [ ; \right ] $ means concatenation. Finally, the stability probability is predicted as $\hat y_{i,k}^{\text{fix}} = \sigma(\text{MLP}_{\text{cls}}(\mathbf{h}_{o_{(i,k)}}^{\text{final}}))$, where $\sigma$ is the activation function. By leveraging the explicit topological graph, this probability reflects a rigorous assessment of the operation's stability within the complex constraint network rather than a mere heuristic estimation based on local statistics.

\subsection{Critical-Path-Aware Mechanism}\label{MTL}
While the heterogeneous graph network $\mathbb{M}_{gnn}$ captures topological structures, training solely on binary stability classification treats all operations homogeneously, ignoring the asymmetric cost of decision errors in FJSP. According to the Critical Path Method~(CPM) principles~\cite{kelley1959critical}, misclassifying a critical operation~(where total slack is $0$) directly degrades the makespan, whereas errors on non-critical nodes are often absorbed by temporal flexibility. To address this, we introduce a critical-path-aware mechanism, $\mathbb{M}_{cpa}$, that adds an auxiliary critical-path-aware task during training. This mechanism injects a bottleneck-oriented inductive bias into the latent space, which explicitly guides the model to prioritise representations that distinguish high-sensitivity bottlenecks from robust operations.

We define operation criticality based on total slack. For an operation $o_{(i,k)}$, the earliest start times $S_{i,k}^E$ and latest start times $S_{i,k}^L$ derive from the disjunctive graph of the current local schedule $\Pi_t$. The slack $S_{i,k}$ is defined as $S_{i,k} = S_{i,k}^L - S_{i,k}^E$.
An operation is labeled critical if $S_{i,k} = 0$, indicating that it lies on the longest path for which any delay strictly increases the makespan.

% \textbf{Dual-Head Architecture.} As is shown in Fig.~\ref{fig:framework}, to inject the training objective, we employ a dual-head architecture sharing the structure-aware embeddings. The main head $\text{MLP}_{\text{fix}}$ predicts the stability probability $\hat{y}_{i,k}^{\text{fix}}$, while the auxiliary head $\text{MLP}_{\text{crit}}$ predicts the criticality probability $\hat{y}_{i,k}^{\text{crit}}$.

\textbf{Supervision Data Collection.} To support the training, we generate supervision signals from two distinct views: \textbf{i)} operation fix labels~$y^{\text{fix}}$: We strictly adhere to the self-labeling protocol defined in L-RHO~\cite{li2025learning}, where an operation is labeled stable $y_{i,k}^{\text{fix}}=1$ if its machine assignment remains consistent with a lookahead oracle schedule. \textbf{ii)} criticality labels~$y^{\text{crit}}$: In contrast, we derive auxiliary labels by analyzing the topological properties of the default schedule $\Pi_{t}^{\text{default}}$. By executing forward and backward passes on the disjunctive graph to compute the total slack, we assign $y_{i,k}^{\text{crit}}=1$ if $S_{i,k} < \epsilon$ indicating the operation lies on the longest path, and $y_{i,k}^{\text{crit}}=0$ otherwise.

\textbf{Dual-Head Architecture.} As is shown in Fig.~\ref{fig:framework}, to inject the training objective, we employ a dual-head architecture sharing the structure-aware embeddings. The main head $\text{MLP}_{\text{fix}}$ predicts the stability probability $\hat{y}_{i,k}^{\text{fix}}$, while the auxiliary head $\text{MLP}_{\text{crit}}$ predicts the criticality probability $\hat{y}_{i,k}^{\text{crit}}$.

\textbf{Training Objective.} The training objective combines the operation fix loss $\mathcal{L}_{\text{fix}}$ and the criticality loss $\mathcal{L}_{\text{crit}}$. Both are formulated as binary cross-entropy losses over the overlap operations $\mathcal{O}_{t}^{overlap}$:
\begin{align}
\mathcal{L}_{\text{fix}} &= - \frac{1}{|\mathcal{O}_{t}^{overlap}|} \sum_{o_{(i,k)}} \left[ y_{i,k}^{\text{fix}} \log \hat{y}_{i,k}^{\text{fix}} + (1 - y_{i,k}^{\text{fix}}) \log (1 - \hat{y}_{i,k}^{\text{fix}}) \right] \\
\mathcal{L}_{\text{crit}} &= - \frac{1}{|\mathcal{O}_{t}^{overlap}|} \sum_{o_{(i,k)}} \left[ y_{i,k}^{\text{crit}} \log \hat{y}_{i,k}^{\text{crit}} + (1 - y_{i,k}^{\text{crit}}) \log (1 - \hat{y}_{i,k}^{\text{crit}}) \right]
\end{align}
Here, $\hat{y}_{i,k}$ is the predicted probability and $y_{i,k}$ is the label. The final joint loss is defined as:
\begin{align}
\mathcal{L}_{\text{total}} = \mathcal{L}_{\text{fix}} + \lambda \mathcal{L}_{\text{crit}}
\end{align}
where $\lambda$ balances the tasks. Optimizing $\mathcal{L}_{\text{crit}}$ acts as a regularizer, penalizing the encoder for discarding topological information about the critical path, thereby ensuring prudent decisions at high-sensitivity nodes.

\subsection{Adaptive Thresholding Strategy}\label{Adaptive}

\begin{figure}
    \centering
    \includegraphics[width=1\linewidth]{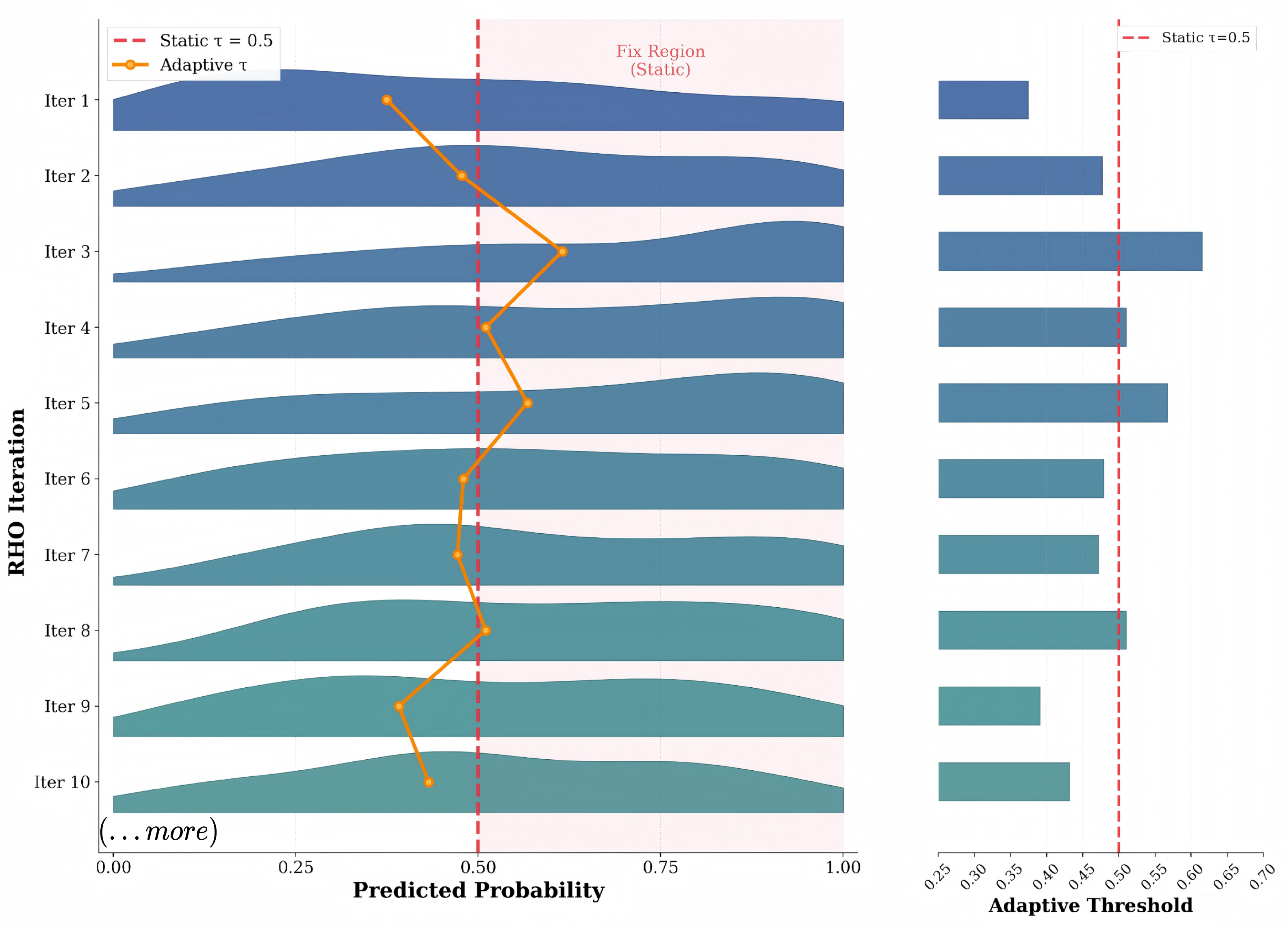}
    \caption{\textbf{Temporal Distribution Shift and Thresholding Strategies.}\textbf{~(Left)} The Ridgeline plot illustrates the evolving density of predicted fix probabilities across RHO iterations. A static threshold~(red line) fails to adapt, whereas our adaptive threshold~(orange line) dynamically tracks the distribution's tail.\textbf{~(Right)} The resulting adaptive threshold values fluctuate to maintain a consistent fixing ratio.}
    \label{fig:ridgeline}
\end{figure}

\begin{table*}[tb]
\centering
\caption{FJSP under Makespan Objective. Each column corresponds to a different FJSP size given by the format: total number of operations, followed by a tuple of ($N_{m}, N_{j}, N_{ops/job}$), i.e., number of machines $|\mathcal{M}|$, jobs $|\mathcal{J}|$, and operations per job. ``Transfer'' means transfer from the 1200 case without fine-tuning.}
\label{tab:fjsp_results}
\resizebox{\textwidth}{!}{% 自动缩放表格以适应页面宽度
\begin{tabular}{lcccccccc}
\toprule
 & \multicolumn{2}{c}{\textbf{600} (10, 20, 30)} & \multicolumn{2}{c}{\textbf{800} (10, 20, 40)} & \multicolumn{2}{c}{\textbf{1200} (10, 20, 60)} & \multicolumn{2}{c}{\textbf{2000} (10, 20, 100), Transfer} \\
\cmidrule(lr){2-3} \cmidrule(lr){4-5} \cmidrule(lr){6-7} \cmidrule(lr){8-9}
 & Time (s) $\downarrow$ & Makespan $\downarrow$ & Time (s) $\downarrow$ & Makespan $\downarrow$ & Time (s) $\downarrow$ & Makespan $\downarrow$ & Time (s) $\downarrow$ & Makespan $\downarrow$ \\
\midrule
CP-SAT (10 hours)~\cite{google_cpsat} & 36000 & 1583 $\pm$ 65 & 36000 & 2128 $\pm$ 75 & 36000 & 3206 $\pm$ 87 & 36000 & 18821 $\pm$ 1986 \\
CP-SAT (30 minutes)~\cite{google_cpsat} & 1800 & 2274 $\pm$ 147 & 1800 & 4017 $\pm$ 413 & 1800 & 10925 $\pm$ 1013 & 1800 & 39585 $\pm$ 2707 \\
GA~\cite{li2016effective} & 1800 & 3659 $\pm$ 87 & 1800 & 5150 $\pm$ 92 & 1800 & 8086 $\pm$ 111 & 1800 & 10406 $\pm$ 640 \\
\midrule
ARD-LNS (Time-based)~\cite{pacino2011large} & 300 & 2100 $\pm$ 269 & 400 & 3298 $\pm$ 365 & 600 & 5115 $\pm$ 558 & 1000 & 14258 $\pm$ 1522 \\
ARD-LNS (Machine-based)~\cite{pacino2011large} & 300 & 3974 $\pm$ 633 & 400 & 6594 $\pm$ 1227 & 600 & 12764 $\pm$ 4206 & 1000 & 52225 $\pm$ 1793 \\
Oracle-LNS (Time-based)~\cite{huang2022anytime} & 300 & 1789 $\pm$ 104 & 400 & 2663 $\pm$ 120 & 600 & 4470 $\pm$ 151 & 1000 & 7275 $\pm$ 785 \\
\midrule
DRL-Echeverria, Greedy~\cite{echeverria2023solving} & 84 $\pm$ 27 & 2347 $\pm$ 189 & 115 $\pm$ 32 & 3105 $\pm$ 161 & 105 $\pm$ 7 & 4570 $\pm$ 218 & 213 $\pm$ 10 & 7673 $\pm$ 356 \\
DRL-Echeverria, Sampling~\cite{echeverria2023solving} & 157 $\pm$ 8 & 2351 $\pm$ 139 & 217 $\pm$ 11 & 3131 $\pm$ 165 & 427 $\pm$ 20 & 4657 $\pm$ 192 & 603 $\pm$ 41 & 7774 $\pm$ 271 \\
DRL-Ho, Greedy~\cite{ho2024residual} & 4 $\pm$ 2 & 3030 $\pm$ 69 & 6 $\pm$ 4 & 4038 $\pm$ 87 & 11 $\pm$ 6 & 6045 $\pm$ 99 & 22 $\pm$ 3 & 10044 $\pm$ 115 \\
DRL-Ho, Sampling~\cite{ho2024residual} & 174 $\pm$ 1 & 2907 $\pm$ 46 & 168 $\pm$ 14 & 3907 $\pm$ 56 & 385 $\pm$ 12 & 5870 $\pm$ 71 & 517 $\pm$ 10 & 9848 $\pm$ 92 \\
DRL-20K, Greedy~\cite{wang2023flexible} & 4 $\pm$ 0.02 & 1628 $\pm$ 72 & 6 $\pm$ 0.04 & 2128 $\pm$ 80 & 10 $\pm$ 0.1 & 3141 $\pm$ 97 & 16 $\pm$ 0.2 & 5184 $\pm$ 114 \\
DRL-20K, Sample 100~\cite{wang2023flexible} & 29 $\pm$ 0.3 & 1551 $\pm$ 60 & 47 $\pm$ 1 & 2048 $\pm$ 69 & 110 $\pm$ 3 & 3063 $\pm$ 81 & 302 $\pm$ 10 & 5082 $\pm$ 98 \\
DRL-20K, Sample 500~\cite{wang2023flexible} & 146 $\pm$ 5 & 1537 $\pm$ 61 & 261 $\pm$ 8 & 2031 $\pm$ 68 & 597 $\pm$ 16 & 3045 $\pm$ 79 & 1738 $\pm$ 16 & 5062 $\pm$ 98 \\
\midrule
Default RHO (Long)~\cite{glomb2022rolling} & 599 $\pm$ 55 & 1529 $\pm$ 58 & 728 $\pm$ 98 & 2044 $\pm$ 75 & 1099 $\pm$ 108 & 3002 $\pm$ 87 & 2871 $\pm$ 244 & 4994 $\pm$ 114 \\
Default RHO~\cite{glomb2022rolling} & 244 $\pm$ 21 & 1558 $\pm$ 73 & 348 $\pm$ 26 & 2103 $\pm$ 78 & 545 $\pm$ 36 & 3136 $\pm$ 91 & 862 $\pm$ 42 & 5207 $\pm$ 114 \\
Warm Start RHO~\cite{glomb2022rolling} & 203 $\pm$ 23 & 1521 $\pm$ 67 & 278 $\pm$ 22 & 2055 $\pm$ 75 & 420 $\pm$ 33 & 3081 $\pm$ 96 & 716 $\pm$ 41 & 5057 $\pm$ 106 \\
% \midrule % 虚线分割，如需实线请改为 \midrule
L-RHO~\cite{li2025learning} & 126 $\pm$ 19 & 1513 $\pm$ 70 & 160 $\pm$ 23 & 2015 $\pm$ 86 & 259 $\pm$ 37 & 3011 $\pm$ 106 & 473 $\pm$ 52 & 4982 $\pm$ 132 \\
\textbf{Graph-RHO~(Ours)} & \textbf{98 $\pm$ 21} & \textbf{1493 $\pm$ 73} & \textbf{139 $\pm$ 29} & \textbf{1985 $\pm$ 80} & \textbf{189 $\pm$ 29} & \textbf{2972 $\pm$ 97} & \textbf{321 $\pm$ 30} & \textbf{4938 $\pm$ 119} \\
\bottomrule
\end{tabular}%
}
\end{table*}

% Currently, relying on a static probability threshold~(e.g., $\tau=0.5$) to determine variable fixation decouples the prediction model from the downstream solver. This approach overlooks the dynamic uncertainty inherent in the rolling horizon process. As the planning window slides across the scheduling horizon, the solver encounters subproblems with varying difficulty: some windows involve obvious decisions, while others involve high contention. Consequently, the model's confidence levels fluctuate significantly over time, rendering a fixed threshold ineffective.

% As visualized in the Fig.~\ref{fig:ridgeline}, the probability density of stability predictions exhibits significant temporal drift. In some iterations (e.g., $t=1$), the probability mass is concentrated at the lower end, indicating low model confidence; in others, it shifts towards higher confidence. A static threshold (red dashed line) imposes a rigid cutoff on these evolving distributions, failing to navigate the delicate trade-off between optimization quality and solving efficiency. When the distribution shifts left, a static threshold results in insufficient pruning, leaving the solver burdened with an excessive number of free variables, thereby drastically reducing computational efficiency. Conversely, when the distribution shifts right, it results in excessive fixing. This overly constrains the remaining optimization space, depriving the solver of the flexibility needed to refine local schedules, ultimately degrading the final makespan.

Relying on a static probability threshold~(e.g., $\tau=0.5$) overlooks the dynamic uncertainty inherent in the rolling horizon process. As visualized in Fig.~\ref{fig:ridgeline}, the probability density of stability predictions exhibits significant temporal drift across iterations. A static threshold imposes a rigid cutoff on these evolving distributions: when the distribution shifts left, it results in insufficient pruning that burdens the solver; when it shifts right, excessive fixing degrades solution quality.

To bridge this gap, we propose the adaptive thresholding strategy $\mathbb{M}_{thr}$ that treats the model's output as a relative ranking and fixes the top subset of operations determined by a pre-set ratio $\gamma$. For each iteration $t$, we first sort the predicted fix probabilities of $N$ overlapping operations $\mathcal{O}_{t}^{overlap}$ in descending order, denoted as $p_{(1)} \ge p_{(2)} \ge \dots \ge p_{(N)}$. To maintain a target fixation ratio $\gamma$, we define the adaptive threshold $\tau_{t}$ as the $k$-th largest probability value:
\begin{align}
\tau_{t} = p_{(k)}, \quad \text{where } k = \lfloor \gamma \cdot N \rfloor
\end{align}
Here, $\lfloor \cdot \rfloor$ denotes the floor function, and $p_{(k)}$ serves as the dynamic decision boundary. The set of fixed operations is then explicitly given by the top-$k$ candidates: $\mathcal{O}_{t}^{fix} = \{o_{i} \mid p_{i} \ge \tau_{t}\}$. To ensure robustness against uniformly low-confidence states, we further apply a safety floor $\tau^{min}$ as $\tau_{t}^{safe} = \max(\tau_{t}, \tau^{min})$.

Ultimately, this rank-based approach effectively decouples pruning decisions from fluctuations in absolute confidence. By consistently fixing the most reliable top-ranked subset defined by the ratio, Graph-RHO guarantees a stable reduction of the combinatorial search space, thereby maintaining a robust equilibrium between solving efficiency and solution feasibility throughout the rolling process.

\section{Experiment}
In this section, we conduct comprehensive empirical evaluations to validate the effectiveness and generalization capability of our Graph-RHO framework. Our experiments are designed to answer the following core research questions:
\begin{itemize}
    \item \textbf{Main Performance~(Sec.~\ref{sec:main_results}).} Does Graph-RHO outperform baselines in solution quality and efficiency across various long-horizon FJSP settings?
    \item \textbf{Zero-Shot Generalization~(Sec.~\ref{sec:generalization}).} Can the model generalize to larger scales (\textbf{Scale Generalization}) and higher loads (\textbf{Load Robustness}) without fine-tuning?
    \item \textbf{Ablation Studies~(Sec.~\ref{sec:ablation}).} What are the contributions of the heterogeneous graph network $\mathbb{M}_{gnn}$, critical-path-aware mechanism $\mathbb{M}_{cpa}$, and adaptive thresholding strategy $\mathbb{M}_{thr}$?
\end{itemize}

% We begin by describing the experimental setup, followed by a detailed analysis of the results for each research question.

\begin{table*}[tb]
    \centering
    \caption{Comparison of generalization capabilities across unseen problem scales and load densities. The model is trained solely on the small-scale ($10, 20, 30$) setting and directly evaluated on larger or denser instances without fine-tuning.}
    \label{tab:generalization}
    \resizebox{\textwidth}{!}{%
    \begin{tabular}{lccccccccc}
        \toprule
        & \multicolumn{4}{c}{\textbf{Scale Generalization}} & & \multicolumn{4}{c}{\textbf{Load Robustness}} \\
        \cmidrule(lr){2-5} \cmidrule(lr){7-10}
        & \multicolumn{2}{c}{$(15, 30, 30)$} & \multicolumn{2}{c}{$(20, 40, 30)$} & & \multicolumn{2}{c}{$(10, 30, 30)$} & \multicolumn{2}{c}{$(10, 40, 30)$} \\
        \cmidrule(lr){2-3} \cmidrule(lr){4-5} \cmidrule(lr){7-8} \cmidrule(lr){9-10}
        Method & Time (s) $\downarrow$ & Makespan $\downarrow$ & Time (s) $\downarrow$ & Makespan $\downarrow$ & & Time (s) $\downarrow$ & Makespan $\downarrow$ & Time (s) $\downarrow$ & Makespan $\downarrow$ \\
        \midrule
        Default RHO & $179 \pm 19$ & $1455 \pm 78$ & $166 \pm 9$ & $1539 \pm 84$ & & $354 \pm 25$ & $2176 \pm 86$ & $471 \pm 30$ & $2864 \pm 85$ \\
        L-RHO & $72 \pm 6$ & $1451 \pm 69$ & $76 \pm 7$ & $1529 \pm 78$ & & $196 \pm 27$ & $2162 \pm 88$ & $213 \pm 36$ & $2860 \pm 105$ \\
        % 下面这一行将 \bm 替换为了 \boldsymbol，通常只需要 amsmath 包
        \textbf{Graph-RHO (Ours)} & \textbf{42 $\pm$ 4} & \textbf{1435 $\pm$ 67} & \textbf{46 $\pm$ 3} & \textbf{1453 $\pm$ 75} & & \textbf{142 $\pm$ 19} & \textbf{2144 $\pm$ 89} & \textbf{160 $\pm$ 20} & \textbf{2831 $\pm$ 101} \\
        \bottomrule
    \end{tabular}%
    }
\vspace{-3pt}
\end{table*}

\subsection{Implementation}
The Heterogeneous GNN encoder is configured with a hidden dimension of $d=64$ and a depth of $L=2$ layers, with $4$ attention heads per GAT. We employ sigmoid activation functions and apply a dropout rate of $0.1$ to mitigate overfitting. The model is trained using the AdamW optimizer with a batch size of $64$ for $200$ epochs. The learning rate is initialized at $1 \times 10^{-4}$ and decayed using a cosine annealing scheduler. To balance the two training objectives, the weight assigned to the auxiliary critical-path-aware task is set to $\lambda=0.5$.

For the RHO setup, we fix the planning window size at $w=80$ and the execution step size at $s=30$. During inference, the confidence-adaptive thresholding strategy employs a target fixation ratio of $\gamma=0.6$, subject to a safety floor of $\tau^{\min}=0.3$. The subproblems are solved using the OR-Tools CP-SAT solver~\cite{google_cpsat}, configured with a linearization level of $2$ to enhance constraint propagation efficiency.

\subsection{Main Results on FJSP with Makespan Objective} \label{sec:main_results}

\textbf{Experimental Setup and Baselines.} We adhere to the protocol established in L-RHO~\cite{li2025learning}, benchmarking on FJSP~(makespan) using distributions from DANIEL~\cite{wang2023flexible} extended to significantly larger horizons. We evaluate three standard problem sizes ($N_{m}, N_{j}, N_{\text{ops/job}}$) with configurations of ($10, 20, 30/40/60$), totaling 600--1,200 operations. Following this protocol, L-RHO and our Graph-RHO are both trained on a dataset of 450 instances
% ~(indicated by the suffix 450 in TABLE~\ref{tab:fjsp_results}) 
and evaluated on 100 test instances. 
Additionally, we evaluate zero-shot transferability on a large-scale setting $(10, 20, 100)$, comprising 2,000 operations, using the model trained on 1,200-operation scale.

Graph-RHO is compared against four baseline categories: 1) \textbf{Global Solvers:} The exact solver CP-SAT~(30 min/10 hr) and the meta-heuristic GA~\cite{li2016effective}; 2) \textbf{Decomposition Heuristics:} ARD-LNS~(Time/Machine-based)~\cite{pacino2011large} and Oracle-LNS~(Time-based); 3) \textbf{Constructive DRL:} State-of-the-art solvers including DRL-Echeverria~\cite{echeverria2023solving}, DRL-Ho~\cite{ho2024residual}, and DRL-20K~\cite{wang2023flexible}~(Greedy/Sampling); and 4) \textbf{RHO Methods:} Default, Warm Start, and the previous SOTA L-RHO~\cite{li2025learning}.

\textbf{Results and Analysis.} TABLE~\ref{tab:fjsp_results} summarizes the performance. Graph-RHO establishes a superior frontier between solution quality and computational efficiency. While constructive DRL baselines~(e.g., DRL-Ho and DRL-20K in Greedy mode) achieve the lowest latency, they suffer from severe quality degradation, yielding makespans that are significantly worse than RHO-based methods~(e.g., DRL-Ho's makespan is nearly double that of Graph-RHO on 2,000-operation scale). In contrast, Graph-RHO delivers state-of-the-art solution quality that rivals or surpasses heavy iterative solvers~(like CP-SAT 10h) on large instances, while remaining orders of magnitude faster. This validates the fundamental advantage of the rolling horizon decomposition for long-horizon scheduling.

Crucially, our Graph-RHO significantly outperforms Default RHO and the previous SOTA L-RHO in both efficiency and quality. Graph-RHO achieves substantial speedups, particularly in the zero-shot transfer setting~($10, 20, 100$). It reduces solve time by 32.1\% compared to L-RHO (473s $\to$ 321s). This efficiency gain is attributed to the heterogeneous GNN encoder, which captures topological constraints more effectively than the MLP used in L-RHO. The structure-aware embeddings enable more precise stability predictions on unseen large-scale graphs, allowing the solver to prune the search space more aggressively without risking feasibility. Moreover, Graph-RHO consistently yields lower makespans than L-RHO~(e.g., 1493 vs. 1513 on 600-operation scale). This quality improvement primarily stems from the critical-path-aware auxiliary task. Unlike standard binary classification objectives, which treat all operations indiscriminately, our model explicitly learns to identify and protect critical operations from being erroneously fixed, ensuring that local pruning decisions do not compromise the global makespan.

\begin{table*}[tb]
    \centering
    \caption{\textbf{Ablation Studies.} We incrementally integrate the three core contributions: Heterogeneous GNN Encoder, Critical-Path-Aware Mechanism, and Adaptive Thresholding, to evaluate their individual impact. $\mathbb{M}_{gnn}$ introduces the GNN Encoder; $\mathbb{M}_{cpa}$ introduces the critical-path-aware objective; $\mathbb{M}_{thr}$ applies the confidence-aware inference strategy.}
    \label{tab:ablation}
    \resizebox{\textwidth}{!}{
    \begin{tabular}{lcccccccccccc}
        \toprule
        & \multicolumn{3}{c}{\textbf{Core Contributions}} & \multicolumn{2}{c}{\textbf{600}} & \multicolumn{2}{c}{\textbf{800}} & \multicolumn{2}{c}{\textbf{1200}} & \multicolumn{2}{c}{\textbf{2000 (Transfer)}} \\
        \cmidrule(lr){2-4} \cmidrule(lr){5-6} \cmidrule(lr){7-8} \cmidrule(lr){9-10} \cmidrule(lr){11-12}
        Method & \small{$\mathbb{M}_{gnn}$} & \small{$\mathbb{M}_{cpa}$} & \small{$\mathbb{M}_{thr}$} & Time (s) $\downarrow$ & Makespan $\downarrow$ & Time (s) $\downarrow$ & Makespan $\downarrow$ & Time (s) $\downarrow$ & Makespan $\downarrow$ & Time (s) $\downarrow$ & Makespan $\downarrow$ \\
        \midrule
        Default RHO~(Baseline) & - & - & - & $244 \pm 21$ & $1558 \pm 73$ & $348 \pm 26$ & $2103 \pm 78$ & $545 \pm 36$ & $3136 \pm 91$ & $862 \pm 42$ & $5207 \pm 114$ \\

        \shortstack[l]{+ $\mathbb{M}_{gnn}$} & $\checkmark$ & - & - & $114 \pm 19$ & $1501 \pm 72$ & $143 \pm 19$ & $1995 \pm 83$ & $250 \pm 36$ & $2987 \pm 99$ & $415 \pm 48$ & $4962 \pm 129$ \\
        \shortstack[l]{+ $\mathbb{M}_{gnn}$ + $\mathbb{M}_{cpa}$} & $\checkmark$ & $\checkmark$ & - & $114 \pm 22$ & $1498 \pm 75$ & $148 \pm 21$ & $1991 \pm 76$ & $232 \pm 29$ & $2982 \pm 103$ & $388 \pm 36$ & $4951 \pm 124$ \\
        \textbf{Graph-RHO (Full)} & $\checkmark$ & $\checkmark$ & $\checkmark$ & $\mathbf{99 \pm 21}$ & $\mathbf{1493 \pm 73}$ & $\mathbf{139 \pm 29}$ & $\mathbf{1985 \pm 80}$ & $\mathbf{189 \pm 29}$ & $\mathbf{2972 \pm 97}$ & $\mathbf{321 \pm 30}$ & $\mathbf{4938 \pm 119}$ \\
        \bottomrule
    \end{tabular}
    }
\vspace{-3pt}
\end{table*}

\subsection{Zero-shot Generation Test Results} \label{sec:generalization}
\textbf{Experimental Setup.} We evaluate zero-shot transferability by applying models trained exclusively on small-scale~($10, 20, 30$) instances directly to unseen scenarios without fine-tuning. We define two distinct protocols: \textbf{i)} \textbf{Scale Generalization:} We expand the problem dimensions to~($15, 30, 30$) and~($20, 40, 30$) while maintaining a constant job-to-machine ratio~($N_j/N_m=2$), thereby testing adaptability to graph expansion. \textbf{ii)} \textbf{Load Robustness:} We intensify resource contention by increasing job counts to 30 and 40 while fixing machine resources~($N_m=10$), effectively elevating the load ratio to $3.0$ and $4.0$.

\textbf{Results and Analysis.} As is shown in TABLE~\ref{tab:generalization}, Graph-RHO demonstrates exceptional zero-shot robustness, consistently outperforming both default RHO and L-RHO across all transfer settings. In the scale generalization test, we observe a clear ``generalization collapse'' in L-RHO. On the source distribution~($10, 20, 30$)~(see in TABLE~\ref{tab:fjsp_results}), L-RHO improves the makespan by 2.9\% over default RHO. However, on the target~($20, 40, 30$) instance, this advantage shrinks to a negligible 0.6\%~($1539 \to 1529$), indicating its learned policy fails to scale. In contrast, Graph-RHO not only maintains but amplifies its advantage, achieving a 5.6\% makespan improvement~($1539 \to 1453$) on the large-scale target. It also reduces inference time by 39.5\% compared to L-RHO~($76s \to 46s$), proving that the heterogeneous GNN encoder captures scale-invariant topological rules rather than overfitting to specific problem sizes. In the load robustness test, Graph-RHO exhibits superior resilience to congestion. As the job-to-machine ratio doubles~($2.0 \to 4.0$) in the~($10, 40, 30$) setting, the default solver struggles. Our Graph-RHO maintains high efficiency~($160s$), delivering a 66\% speedup over default RHO~($471s$). More importantly, it widens the quality gap against L-RHO, reducing the makespan by 29 units ($2860 \to 2831$), whereas L-RHO barely outperforms the default baseline in this high-contention regime. The superior generalization stems from our GNN network's ability to learn scale-invariant topological rules, unlike L-RHO's reliance on statistical aggregations that succumb to distribution shifts. By capturing relative structural roles rather than absolute values, the learned policy seamlessly adapts to larger or denser graphs.

\subsection{Ablation Studies}\label{sec:ablation}
We perform a progressive ablation study to isolate the impact of our three core contributions: the heterogeneous GNN encoder $\mathbb{M}_{gnn}$, the critical-path-aware mechanism $\mathbb{M}_{cpa}$, and the adaptive thresholding strategy $\mathbb{M}_{thr}$. We incrementally integrate these components into the default RHO baseline to assess their additive benefits. 
As shown in Table~\ref{tab:ablation}, we observe a clear cumulative performance gain across all problem scales. The introduction of the $\mathbb{M}_{gnn}$~(Row 2) consistently accelerates inference and improves the solution quality compared to the baseline (e.g., 59\% time reductions and 5\% makespan reductions on 800-operation scale), confirming that structural encoding is fundamental for topological feature extraction. Adding the critical-path-aware mechanism $\mathbb{M}_{cpa}$ (Row 3) acts as a robust stabilizer, yielding consistent makespan reductions (e.g., 6.5\% time reduction and 0.2\% makespan reduction on 2,000-operation scale) by preventing the model from erroneously fixing bottleneck operations. Moreover, integrating adaptive thresholding $\mathbb{M}_{thr}$ into the full Graph-RHO model delivers the largest efficiency boost, slashing solve time by over 17\% on the largest 2,000-operation transfer task compared to the static thresholding variant, without compromising solution quality. This validates that the synergy of topological reasoning, critical-path-aware mechanism, and adaptive thresholding inference is essential for pushing the frontier of long-horizon FJSP scheduling.

\section{Conclusion}
In this work, we propose Graph-RHO, a critical-path-aware graph-based RHO framework for long-horizon FJSP. By synergizing a topology-aware heterogeneous graph encoder with edge-feature-aware message passing, a critical-path-aware auxiliary training objective, and an adaptive thresholding inference strategy, Graph-RHO establishes a new state of the art in both solution quality and computational efficiency across various FJSP settings, while exhibiting strong zero-shot generalization on large unseen instances without fine-tuning.

\bibliographystyle{IEEEtran}
\bibliography{references}

\vspace{12pt}
% \color{red}
% IEEE conference templates contain guidance text for composing and formatting conference papers. Please ensure that all template text is removed from your conference paper prior to submission to the conference. Failure to remove the template text from your paper may result in your paper not being published.

\end{document}